\documentclass[letterpaper, 10pt, conference]{ieeeconf}

\IEEEoverridecommandlockouts

\usepackage{times}
\usepackage{graphicx}
\usepackage{bm}
\usepackage{booktabs}
\usepackage{amsmath}
\usepackage{amssymb}
\usepackage{color}
\usepackage[short]{optidef}
\usepackage{gensymb}
\usepackage{multirow}
\usepackage[mathscr]{eucal}
\usepackage{xparse,soul}
\usepackage[flushleft]{threeparttable}
\usepackage{makecell}
\usepackage{fancyhdr}
\usepackage{cite}
\usepackage{algorithm}
\usepackage{bm}
\usepackage{algorithm}
\usepackage{algpseudocode}
\usepackage{float}
\usepackage{float}

\usepackage{array}
\newcolumntype{P}[1]{>{\centering\arraybackslash}p{#1}}

\title{A Learning-Based Approach for  Contact Detection, Localization, and Force Estimation   of Continuum Manipulators With  Integrated OFDR Optical Fiber}
\author{Mobina Tavangarifard$^{1}$, Jonathan S.~Kacines$^{*1}$, Qiyu Li$^{*2}$, and Farshid Alambeigi$^{1}$
\thanks{*These authors contributed equally to this work.}
\thanks{This work was supported by the Collaborative Accelerator for Transformative Research Endeavors grant, jointly awarded by The University of Texas at Austin and The University of Texas MD Anderson Cancer Center.}
\thanks{$^{1}$M.~Tavangarifard, J.~S.~Kacines, and F.~Alambeigi are with the Walker Department of Mechanical Engineering and Texas Robotics, The University of Texas at Austin, Austin, TX 78712, USA. Email: \{mt39884, jsk2852\}@my.utexas.edu, farshid.alambeigi@austin.utexas.edu.}
\thanks{$^{2}$Q.~Li is with the Department of Computer Science, The University of Texas at Austin, Austin, TX 78712, USA. Email: ql4239@my.utexas.edu.}
}
\begin{document}
\maketitle
\pagestyle{empty}

\begin{abstract}
Continuum manipulators (CMs) are widely used in minimally invasive procedures due to their compliant structure and ability to navigate deep and confined anatomical environments. However, their distributed deformation makes force sensing, contact detection, localization, and force estimation challenging, particularly when interactions occur at unknown arc-length locations along the robot.
To address this problem, we propose a cascade learning-based framework (CLF) for CMs instrumented with a single distributed Optical Frequency Domain Reflectometry (OFDR) fiber embedded along one side of the robot. The OFDR sensor provides dense strain measurements along the manipulator backbone, capturing strain perturbations caused by external interactions.
The proposed CLF first detects contact using a Gradient Boosting classifier and then estimates contact location and interaction force magnitude using a CNN--FiLM model that predicts a spatial force distribution along the manipulator. Experimental validation on a sensorized tendon-driven CM in an obstructed environment demonstrates that a single distributed OFDR fiber provides sufficient information to jointly infer contact occurrence, location, and force in continuum manipulators.

\end{abstract}



\section{Introduction}

Thanks to their compliant structure, continuum manipulators (CM) can follow tortuous paths, maneuver through narrow channels, and reach deep-seated targets that are inaccessible to conventional rigid manipulators\cite{Webster2009TubeContinummRobot}. In particular, in minimally invasive and natural orifice transluminal endoscopic procedures, this flexibility reduces tissue trauma, enhances access to complex anatomical regions, and improves procedural safety \cite{Shaikh2010NOTESReview}.
However, this same characteristic introduces significant challenges in both shape and force estimation, particularly when these robots operate in constrained anatomical environments where external contact can alter their expected configuration\cite{Ferguson2024UnifiedShape}. Interactions may occur at unknown locations along the structure when moving inside the body, generating point or distributed contact forces, thereby complicating accurate contact detection, localization, and force estimation \cite{Hu2025ForceEstimationReview}.

\begin{figure}[h!]
    \centering
    \includegraphics[width = \columnwidth]{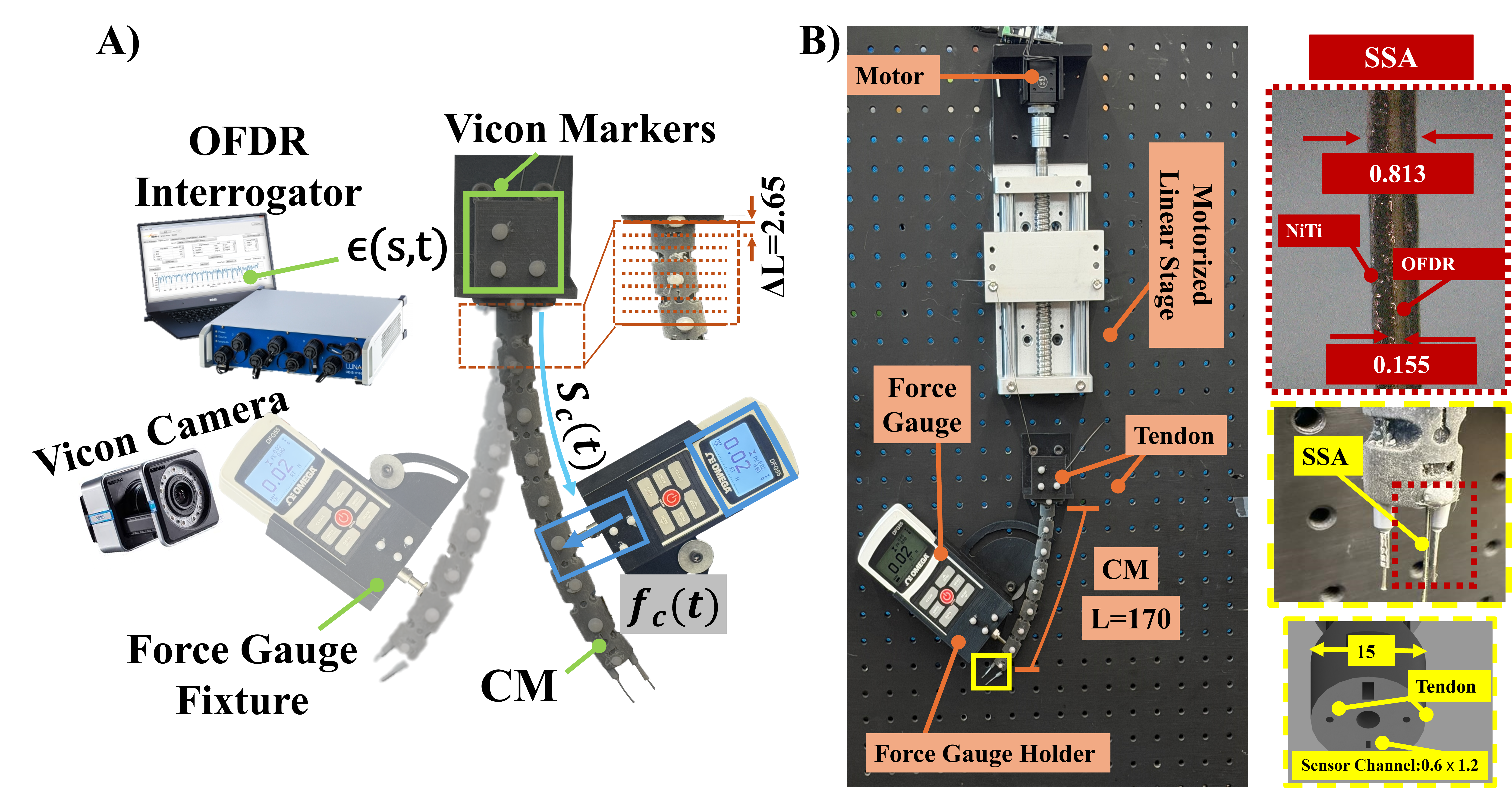}
    \caption{A) Conceptual illustration of a continuum manipulator interacting with an obstacle at an unknown location along its geometry. The interaction force and contact location are measured using a force gauge and a motion capture system. An embedded OFDR optical fiber measures the distributed strain along the manipulator.
B) Physical experimental setup used to evaluate the proposed CLF. The figure shows the shape-sensing assembly and its integration inside the continuum manipulator (CM).}
    \label{concept}
\end{figure}

To address the aforementioned challenges and difficulties in force estimation of CMs, the literature has generally explored three primary categories of approaches: Model-based methods\cite{BACK20183DForceEstimation}, Image-based approaches\cite{chua2022forceestimationrobotassistedsurgery}\cite{Jung2021VisionBasedEstimation}, and integrated sensors. \textit{Model-based methods} typically employ beam theory, Cosserat rod formulations, or piecewise constant curvature assumptions combined with observers to infer external loads from discrepancies between predicted and measured deformation \cite{Walker2023ContinuumOverview} \cite{Liu2022DistributedForce}. While these approaches integrate naturally with control frameworks, their accuracy depends heavily on parameter identification and modeling assumptions, particularly under frictional contact and complex boundary conditions. 
\textit{Image-based approaches} attempt to estimate interaction forces from visual deformation cues using fluoroscopy, endoscopy\cite{Liu2022Endoscopy}, or learning-based pixel-to-force mappings\cite{Neidhardt2023OpticalForce}. Although they avoid embedded sensors, their robustness is limited in occluded and contact-rich surgical scenes, and fluoroscopic systems impose radiation constraints\cite{narain2017radiation}.

\textit{Embedded sensing} for force estimation in continuum robots can generally be categorized into three main types: (1) load-cell–based sensing\cite{zhang2025integratedshapeforceestimationcontinuum}, (2) magnetic sensing\cite{Abah2022MultiModalSensorArray}, and (3) optical fiber–based sensing, including Fiber Bragg Grating (FBG) and Optical Frequency Domain Reflectometry (OFDR)\cite{Tavangarifard2024SingleFiberOFDR}. Each approach offers distinct advantages and limitations in terms of integration complexity, observability, robustness, and clinical suitability.
Load-cell–based sensing is the most direct method for force measurement and is typically implemented at the base, tip, or along actuation tendons of the robot. These sensors provide accurate and high-bandwidth force readings and are relatively simple to calibrate. However, aside from difficult integration with CMs and their rigid structure, they generally measure forces only at discrete locations and cannot directly localize distributed contacts along the backbone\cite{zhang2025integratedshapeforceestimationcontinuum}.

\begin{figure*}[t]
    \centering
    \includegraphics[width = 2\columnwidth]{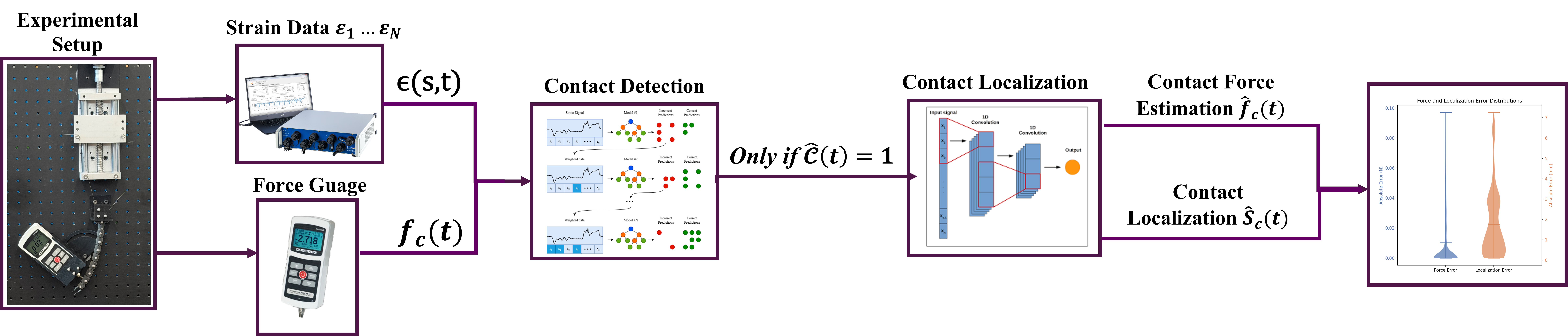}
    \caption{The proposed CLF architecture for contact detection, localization, and magnitude of force estimation.}
    \label{flow chart}
\end{figure*}
Magnetic sensing estimates robot shape and external forces by tracking embedded magnets or magnetic sensors, often combined with Cosserat rod models for 3D shape–force reconstruction \cite{Yousefi2023Modelaided3S}. These systems are compact and line-of-sight independent, making them suitable for minimally invasive procedures. However, they are highly sensitive to magnetic disturbances and ferromagnetic materials common in clinical environments, and they require careful calibration and external field generators or sensor arrays, increasing system complexity and reducing robustness.

Optical fiber–based sensing, particularly FBG sensors, has become one of the most widely adopted embedded sensing strategies for continuum robots. FBGs provide high-resolution strain measurements, can be miniaturized, and are biocompatible. Several works have demonstrated force estimation by mapping discrete FBG strain measurements to external loads through beam or Cosserat modeling\cite{Khan2017FBGForceSensing}. 

Nevertheless, FBG-based systems are limited by discrete sensing points, which restrict spatial resolution and reduce accuracy under distributed contact.  Moreover, fabrication and customization of FBG arrays are costly and technically complex, involving specialized inscription equipment, and careful integration, increasing overall system cost and manufacturing complexity \cite{Alambeigi2020SCADE}.
To address the limitations of sparse sensing, distributed OFDR optical fiber sensing  has emerged as a promising alternative to discrete FBG arrays.  OFDR enables continuous strain measurement with millimeter-scale spatial resolution\cite{Francoeur2024DistributedNeedleOFDR}, providing dense strain fields that capture localized deformation and distributed contact interactions with high fidelity\cite{Monet2020HighResolutionOFDRvsFBG}. This improved spatial observability enhances contact localization and force estimation, particularly in constrained environments with unknown interaction locations.

Rather than explicitly reconstructing curvature and solving inverse mechanics equations in optical fibers, recent works have explored learning-based frameworks that treat strain measurements as structured spatial data and directly infer interaction states \cite{Sincak2024SensingContinuumReview, Wang2021MLContinuumSurvey}. For example, neural networks are trained to map strain features of FBG arrays to contact-related information such as collision detection\cite{Sefati2021DataDrivenShapeSensingFBG}.  However, similar to most learning-based methods that rely on discrete FBG sensing, these approaches suffer from limited spatial resolution and reduced observability, especially under distributed or multiple contacts \cite{Monet2020HighResolutionOFDRvsFBG}. In addition, they are largely restricted to binary contact detection, with limited capability for accurate contact localization or force estimation. While OFDR has been primarily employed for high-resolution shape reconstruction, its potential for learning-based interaction inference remains largely unexplored. 

To collectively address the aforementioned challenges and as our main contributions, this paper leverages a single distributed OFDR optical fiber to simultaneously solve three tightly coupled interaction-aware tasks in continuum manipulators (CMs): contact detection, contact localization, and force estimation. Unlike prior work limited to binary contact classification using sparse FBG sensing, we exploit the dense strain field provided by OFDR to capture rich, spatially continuous strain patterns along the entire manipulator.
We then introduce a Cascade Learning Framework (CLF) that first detects the presence of contact and subsequently estimates both the location and magnitude of the external interaction. The proposed CLF is experimentally validated on a CM sensorized with a single OFDR fiber interacting with obstacles along its structure.


\section{Problem Definition and System Overview }

\subsection{Problem Definition}

As shown in Fig. \ref{concept}, let's consider a tendon-driven CM sensorized with a \textit{single} distributed OFDR fiber embedded along \textit{one side} of its body. The manipulator bends dynamically in an environment containing unknown obstacles, and contact may occur at an unknown location along its structure. The manipulator shape, contact type, and contact location are not known \textit{a priori}. The only available sensing modality is the time-varying distributed strain field $\epsilon(s,t)$ provided by the OFDR fiber, where $s \in [0,L]$ denotes the arc-length coordinate along the manipulator of length $L$, and $t \in [0,T]$ represents actuation time.
Given the measured strain distribution along the arc length  of the manipulator during bending, our \textit{objective} is to determine: (1) Contact state $\mathcal{C}(t) \in \{0,1\}$ indicating whether external contact has occurred; (2) Contact location $s_c(t) \in [0,L]$ representing the arc-length position of contact along the manipulator;
 and (3) Contact force magnitude $F_c(t) \in \mathbb{R}^{+}$ corresponding to the external interaction force. To solve this multi-stage inference problem, we propose a Cascade Learning Framework (CLF) that  defines a mapping $\mathcal{F}: \epsilon(s,t) \longrightarrow \{\mathcal{C}(t),\, s_c(t),\, F_c(t)\}$. The CLF first performs contact detection and, conditioned on $\mathcal{C}(t)=1$, subsequently estimates $s_c(t)$ and $F_c(t)$ from the  strain measurements.

As shown in Fig. \ref{concept}, the proposed approach is validated using a custom-designed tendon-driven continuum manipulator with an integrated OFDR fiber. The experimental setup includes a load cell mounted on a fixation jig that enables controlled variation of contact angle and location, as well as a motion capture system with markers attached along the manipulator body and load cell mount jig to provide ground-truth shape and contact localization. The following sections describe the experimental protocol and the proposed CLF in detail.

\subsection{Continuum Manipulator Design and Fabrication}
As shown in Fig.~1, in this study we used a tendon-driven continuum manipulator (CM) with symmetric notch patterns, a total length $L = 170$~mm, and circular cross section with outer diameter of 15~mm. Based on this geometry, the CM is divided into $N=64$ identical sections, each with equal length $\Delta L = L/N= 2.65 mm$ shown in Fig. \ref{concept}. Of note, this 2.65 mm resolution is determined by the physical footprint of the force-gauge contact probe. These sections are later used in the proposed CLF to discretize the arc-length domain for contact localization and force estimation. In other words, this corresponds to the minimum spatial resolution used for contact localization.
Two tendons are routed through channels embedded along the left and right sides of the manipulator, parallel to the sensing assembly. The tendons are actuated by a motor-driven ball screw, where torque from the DC motor (Dynamixel XM430) is transmitted to the screw through a flexible helical beam coupling, which enables standardized and repeatable bending motions based on controlled motor displacement. By independently pulling either tendon, lateral bending is generated, steering the CM toward the corresponding direction. The CM was additively manufactured using a Stratasys J750 3D printer with TangoBlack material.

To obtain ground-truth measurements, optical markers are attached along the CM backbone and tracked using the Vicon Inc. Vero motion capture system to reconstruct the true shape of the manipulator and accurately compute the contact location. Three markers are attached to the CM base mount to define a fixed global reference frame. Nine markers are evenly distributed along the manipulator backbone to capture its spatial configuration and reconstruct the ground-truth shape. Further, a dedicated rectangular internal channel ($1.3\text{mm} \times 0.6\text{mm}$) is incorporated along the manipulator length to embed the OFDR shape-sensing assembly. This channel allows continuous strain measurements $\epsilon(s,t)$ along the arc-length coordinate, which serve as the sole sensing input to the proposed CLF for contact detection, localization, and force estimation.

\subsection{Distributed OFDR Shape-Sensing Assembly}

As shown in Fig. \ref{concept}, the CM is sensorized with a single OFDR Shape-Sensing Assembly (OFDR-SSA). The SSA consists of a single optical fiber (HD65, Luna Innovations Inc.) with a diameter of 0.155~mm, bonded along the midline of a $0.152~\text{mm} \times 0.813~\text{mm}$ NiTi wire using cyanoacrylate adhesive. The assembly is fabricated following the procedure described in~\cite{Tavangarifard2024SingleFiberOFDR} and subsequently routed through the aforementioned internal sensor channel embedded along one side of the CM.
The OFDR system provides continuous distributed strain measurements along the full length of the optical fiber. When the CM bends, the embedded fiber undergoes deformation, generating a spatial--temporal strain $\epsilon(s,t), \quad s \in [0,L], \; t \in [0,T]$. Because the fiber is positioned asymmetrically relative to the neutral bending axis as shown in Fig. \ref{concept}, curvature induces measurable strain variations along the arc length during dynamic motion of the CM. When external contact occurs, localized perturbations are introduced into the distributed strain profile, creating distinguishable spatial--temporal signatures that are later used for contact detection, localization, and force estimation.

\subsection{Controlled Contact Generation Setup}

As shown in Fig. \ref{concept}, to simulate obstacle interaction with the CM in a controlled and repeatable manner, a force gauge (Omega DFG55) is introduced as a fixed ``obstacle'' within the bending workspace of the manipulator. This obstacle constrains the CM trajectory and replicates contact conditions encountered in realistic surgical environments.
As the CM bends into the obstacle, the resulting interaction force is recorded by the force gauge, providing ground-truth measurements of the contact force magnitude $F_c(t)$. The gauge is rigidly mounted to an optical table through a custom-designed additively manufactured jig (see Fig. \ref{concept})  using Raise3D 3D printer and PLA material. This fixture enables positioning the force gauge at 8 discrete arc-length locations and adjustable contact angles on both sides of the manipulator. This design allows systematic variation of contact location and orientation consistent with the CM architecture.
The force gauge mount is instrumented with three optical markers tracked by motion capture system to determine its position and orientation in space.
Using these measurements, the contact point location along the arc length is computed precisely, providing ground-truth data for validating the estimated contact location $s_c(t)$. 
Together, the motion capture system and force gauge provide accurate reference measurements for evaluating the CLF’s ability to detect contact $\mathcal{C}(t)$, localize the interaction $s_c(t)$, and estimate the contact force magnitude $F_c(t)$.

The following section describes the data acquisition procedure and the proposed learning framework in detail.


\subsection{Data Acquisition Protocol}
Using the experimental setup shown in Fig.~\ref{concept}, two sets of experiments were conducted to evaluate the efficacy of single-fiber contact detection, localization, and force estimation. Since the SSA has been inserted in one side of CM, experiments were performed on both bending directions to assess performance under tensile and compressive strain regimes. For each set, the force gauge was mounted sequentially on each side of the manipulator, and eight distinct contact locations were defined along the CM body. Each location was tested three times, resulting in repeated measurements across both surfaces.
For each trial, the motor was commanded to slowly pull the actuation cables until contact with the force gauge was established. The load cell measured the contact force, and actuation was stopped once the force reached 0.1~N. This force level ensured safe interaction without inducing local deformation of the CM. After reaching 0.1~N, the motor was returned to its initial position, and the procedure was repeated three times per location. The entire protocol was performed across all eight locations on both sides of the manipulator.
Throughout each trial, distributed strain measurements from the embedded OFDR fiber and synchronized force data were continuously recorded, capturing both pre-contact deformation and post-contact interaction dynamics. We denote the measurements as $\mathrm{cv}_{i}$ and $\mathrm{cc}_{i}$, where $i=1,\dots,8$ corresponds to the contact location index. The notation $\mathrm{cv}$ (convex) refers to tensile strain when bending toward the fiber, while $\mathrm{cc}$ (concave) denotes compressive strain when bending away from the fiber.

All measurements were acquired through a ROS~2-based data collection pipeline integrating the Luna ODiSI interrogator (ODiSI 6000 Series, Luna Innovations Inc.), the linear actuator, and the force gauge; the interrogator streams distributed strain data at 1~kHz via the OMSP TCP interface as UTF-8 JSON frames, and a dedicated ROS~2 driver node (\texttt{odisi\_driver}) maintains a persistent socket connection, parses incoming frames, and publishes strain measurements as ROS messages using a newest-only buffering strategy to prevent backlog during high-rate streaming. 
The linear actuator driven by the Dynamixel motor is connected over a USB serial link at $57,000$~baud and the force gauge USB serial connection is at $115,200$~baud. For experiments requiring geometric ground truth, motion-capture measurements were acquired using a Vicon system through the \texttt{ros-vicon-bridge}, and marker poses were synchronized with strain, actuator, and force measurements within the ROS~2 framework. The entire acquisition stack was containerized to ensure repeatability, and during each experiment a logger node recorded synchronized strain vectors, actuator commands, force readings, and (when applicable) marker positions into time-stamped CSV files at $20$~Hz and was terminated automatically when the force reading exceeds a threshold of $0.10$~N. The data collected were subsequently used as inputs to the proposed CLF.

\section{Cascade Learning Framework (CLF) }
The CLF defines the following mapping
\begin{equation}
\mathcal{F}: \epsilon(s,t) \longrightarrow \{\mathcal{C}(t),\, s_c(t),\, F_c(t)\}. 
\end{equation}
In this definition and as shown in Fig. \ref{flow chart}, it first performs contact detection and, conditioned on $\mathcal{C}(t)=1$, subsequently estimates $s_c(t)$ and $F_c(t)$ from the strain measurements. The following sections define the cascaded formulation.  
\subsection{Contact Detection Learning Model}

\subsubsection{Training Data and Contact Detection Formulation} As the first step of CLF, we formulate contact detection problems as a binary classification problem and solve it using supervised machine learning methods. With the OFDR measurements, for each data frame sample $k$ corresponds to a discretized strain profile along the arc length at time $t_k$, with the interval of each time frame being $0.05$~s (i.e., $20$~Hz). We define the measured input strain as
\begin{equation}
\big[\epsilon(s_1,t_k),,\epsilon(s_2,t_k),,\ldots,,\epsilon(s_{263},t_k)\big]^T,
\end{equation}
where $s_i \in [0, L]$ denotes the $i$-th arc-length sampling location, $i \in [0, 263]$, along the embedded section of OFDR fiber. The interrogator samples strain at $0.65$~mm gauge pitch, giving $263$ strain nodes over the length $L = 170$~mm section. A synchronized force-gauge measurement provides the contact force magnitude $F_c(t_k)$.
To generate the ground-truth contact label $\mathcal{C}(t_k)$, we use the force gauge reading, defined as:
\begin{equation}
\label{eq:contact_force}
\mathcal{C}(t_k) \in \{0, 1\}, \qquad
\mathcal{C}(t_k) = [F_c(t_k) \ge F_\text{thresh}]
\end{equation}
In our experiments, we define the threshold $F_\text{thresh}$ as $0.01$~N which in the resolution of the used force-gauge.

\subsubsection{Contact Classification Model}

\begin{figure}[t]
    \centering
    \includegraphics[width = 0.8\columnwidth]{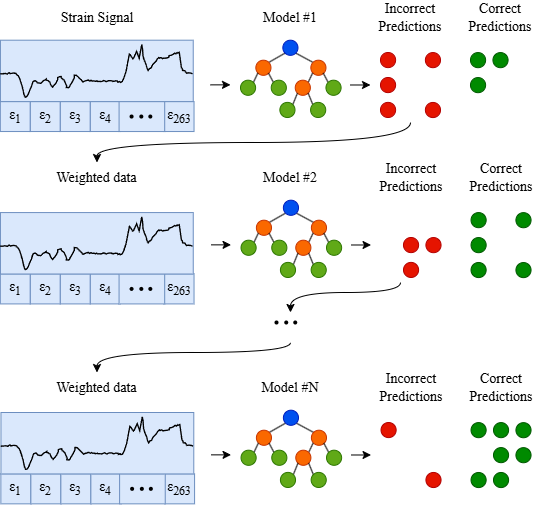}
    \caption{Gradient Boosting Classifier for Contact Detection. Figure illustrates the cascade process using weighted strain data for contact classification.}
    \label{fig:gb_diagram}
\end{figure}

 To reduce the dimensionality of the distributed strain input while retaining its spatial structure, we apply a uniform downsampling transform. We downsample the strain vector by partitioning it into continuous segments and replacing each segment with the average of the strain node values.  The downsample-dimension is a parameter that is tuned for optimal result in our learning pipeline. The resulting feature vector is used as input to a Gradient Boosting decision-tree classifier, depicted in Fig. \ref{fig:gb_diagram}.
The classifier is trained using a predefined split strategy to evaluate generalization across experimental runs. Specifically, we adopt a 
-test-id approach, where \texttt{test\_id} data collected from a unique obstacle location and each fold holds out all samples from one \texttt{test\_id} for testing while training on the remaining runs.

\subsection{Contact Localization and Force Estimation Model}
\subsubsection{Training Data and Model Formulation} As shown in Fig.~\ref{flow chart}, once contact is positively classified by the proposed CLF, a CNN-based model estimates both the magnitude and spatial region of the contact force. To generate training labels and establish ground-truth contact localization, the mentioned multi-camera motion capture system was first used, with nine reflective markers placed along the CM and on the force-gauge mount. The 3D marker coordinates are defined as
\begin{equation}
\label{eq:marker_xyz}
m_i(t) = [x_i(t),\, y_i(t),\, z_i(t)], \quad i=1,\dots,9,
\end{equation}
from which the CM backbone is reconstructed and the ground-truth arc-length contact location $s_c^{gt}(t_k)$ is computed at each time step.
At time $t_k$, the CNN input is the distributed strain vector 
$\bm{\varepsilon}_k = [\epsilon(s_1,t_k), \dots, \epsilon(s_N,t_k)]^\top \in \mathbb{R}^N$ together with the motor position $\delta(t_k)$. As shown in Fig. \ref{concept}, the CM length is discretized into 64 regions with arc-length coordinates $s_j$, and the ground-truth force distribution is encoded as a force-scaled Gaussian
\[
\tau_j^{gt}(t_k) = F_c(t_k)\exp\!\left(-\frac{(s_j - s_c^{gt}(t_k))^2}{2\sigma^2}\right),
\]
where $F_c(t_k)$ is the measured contact force magnitude and $\sigma$ controls spatial spread. The model outputs a predicted force distribution $\bm{\tau}(t_k) = [\tau_1(t_k),\dots,\tau_{64}(t_k)]$, from which the scalar force magnitude and contact region are estimated as $F_c(t_k)=\max_j \tau_j(t_k)$ and $j^*(t_k)=\arg\max_j \tau_j(t_k)$, respectively. Alternatively, a Softmax layer can be applied to obtain a probabilistic contact localization.


\begin{figure}[t]
    \centering
    \includegraphics[width = 1\columnwidth]{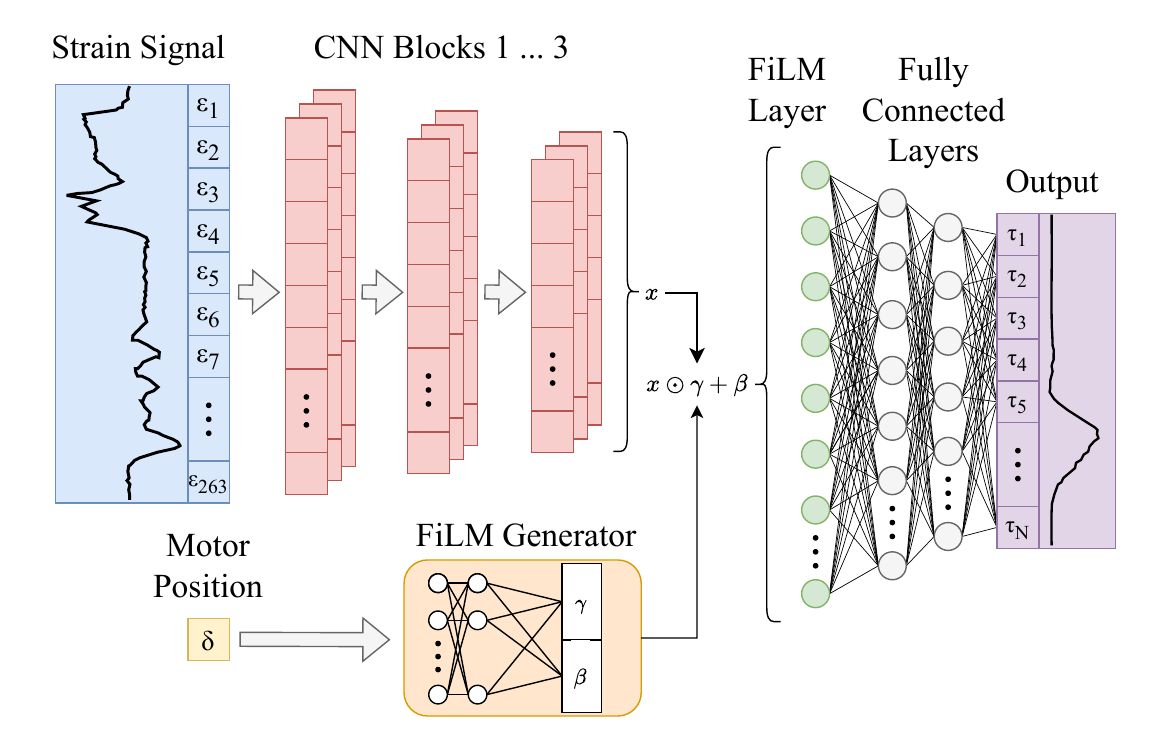}
    \caption{Illustration of the CNN architecture with FiLM generator, processing strain signals and motor position for force localization.}
    \label{cnn}
\end{figure}
\subsubsection{Force Localization and Estimation Model}
\label{section:CNN-Film}
To estimate the magnitude and spatial region of the contact force along the CM, we employ a Convolutional Neural Network (CNN) with a Feature-wise Linear Modulation (FiLM) generator to encode the scalar motor position (see Fig. \ref{cnn}). The inputs to the model are the instantaneous strain vector $\bm{\varepsilon}_k$ and the motor position $\delta(t_k)$. The network defines the forward mapping
\begin{equation}
\hat{\bm{\tau}} = f_{\theta}(\bm{\varepsilon}_k,\, \delta(t_k)),
\end{equation}
where $f_{\theta}(\cdot)$ denotes the CNN--FiLM model parameterized by $\theta$, and $\hat{\bm{\tau}} \in \mathbb{R}^{N_{\text{out}}}$ represents the predicted discrete force distribution along the CM.

The CNN architecture consists of two main components. First, the strain signal $\bm{\varepsilon}_k$ is processed through three 1D convolutional blocks (Conv1d $\rightarrow$ BatchNorm1d $\rightarrow$ ReLU $\rightarrow$ MaxPool1d $\rightarrow$ Dropout) to extract spatial features (see Table~\ref{tab:conv_blocks}). Second, the scalar motor position $\delta(t_k)$ is passed through a FiLM generator composed of two fully connected layers that output modulation parameters $\gamma$ and $\beta$ (see Table~\ref{tab:film_layer}). The convolutional features $x$ are modulated according to
\begin{equation}
\text{FiLM}(x,\gamma,\beta) = \gamma \odot x + \beta,
\label{eq:film}
\end{equation}
where $\odot$ denotes the Hadamard (element-wise) product. The FiLM-modulated features are subsequently processed by three fully connected layers to produce the predicted force distribution $\hat{\bm{\tau}}$ (see Table \ref{tab:fc_layers}). 

During training, the model is supervised using a Gaussian-encoded ground-truth force distribution $\bm{\tau}^{gt}$ derived from motion-capture-based localization and scaled by the measured contact force magnitude. The network parameters $\theta$ are optimized using a mean squared error (MSE) objective:
\begin{equation}
\mathcal{L}_{\mathrm{MSE}}(\theta)
= \frac{1}{N_{\text{out}}} 
\sum_{j=1}^{N_{\text{out}}}
\left(
\hat{\tau}_j - \tau_j^{gt}
\right)^2.
\end{equation}

During inference, the predicted contact force magnitude is obtained as
\begin{equation}
\hat{F}_c(t_k) = \max_{j}\, \hat{\tau}_j,
\end{equation}
and the predicted contact region index is computed as
\begin{equation}
j^* = \arg\max_{j}\, \hat{\tau}_j, 
\qquad
\hat{s}_c(t_k) = s_{j^*}.
\end{equation}

This structured formulation enables simultaneous regression of contact force magnitude and localization through prediction of a spatially distributed force profile.

\begin{table}[t]
\centering
\caption{Convolutional Block Architecture}
\label{tab:conv_blocks}
\begin{tabular}{lccccc}
\hline
\textbf{Block} & \textbf{In Ch.} & \textbf{Out Ch.} & \textbf{Kernel} & \textbf{Padding} & \textbf{Dilation}\\
\hline
Block 1 & 1  & 16 & 7 & 3 & 2\\
Block 2 & 16 & 32 & 7 & 3 & 2\\
Block 3 & 32 & 64 & 3 & 1 & 1\\
\hline
\end{tabular}
\begin{tablenotes}
\small
\item Each block consists of Conv1d $\to$ BatchNorm1d $\to$ ReLU $\to$ MaxPool1d $\to$ Dropout.
\end{tablenotes}
\end{table}

\begin{table}[t]
\centering
\caption{FiLM Generator Architecture}
\label{tab:film_layer}
\begin{tabular}{lccc}
\hline
\textbf{Layer} & \textbf{In Features} & \textbf{Out Features} & \textbf{Purpose} \\
\hline
Linear 1 & $N_{\text{scalar}}$ & 128 & Feature projection \\
Linear 2 & 128 & 128 & Output $\gamma$ and $\beta$ \\
\hline
\end{tabular}
\end{table}

\begin{table}[t]
\centering
\caption{Fully Connected Layer Architecture}
\label{tab:fc_layers}
\begin{tabular}{lccc}
\hline
\textbf{Layer} & \textbf{In Features} & \textbf{Out Features} & \textbf{Activation} \\
\hline
Linear 1 & $Dim_{\text{FiLM}}$ & 128 & ReLU \\
Linear 2 & 128 & 64 & ReLU \\
Linear 3 & 64 & $N_{\text{out}}$ & -- \\
\hline
\end{tabular}
\end{table}


\subsection{Evaluation Metrics}

To comprehensively assess the performance of the CLF model, we used the following evaluation metrics: 



\subsubsection*{Contact Classification Metrics}
Let $\{(\mathcal{C}(t_k), \hat{\mathcal{C}}(t_k))\}$ denote the ground-truth and predicted contact labels, respectively, where both take values in $\{0,1\}$. A True Positive (TP) occurs when both the predicted and ground-truth labels indicate contact, while a True Negative (TN) occurs when both indicate no contact. A False Positive (FP) corresponds to incorrectly predicting contact when none exists, and a False Negative (FN) corresponds to failing to detect an actual contact. 
Precision is defined as the proportion of predicted contact events that are correct (TP divided by TP + FP), while recall is defined as the proportion of actual contact events that are correctly detected (TP divided by TP + FN). The True Positive Rate (TPR) is equivalent to recall, and the False Positive Rate (FPR) is defined as the proportion of non-contact instances incorrectly classified as contact (FP divided by FP + TN).
The Receiver Operating Characteristic (ROC) curve is constructed by plotting the TPR against the FPR as the decision threshold varies. The Area Under the ROC Curve (ROC--AUC) provides a threshold-independent measure of classifier performance, summarizing the trade-off between sensitivity and false alarm rate across all thresholds.


\subsubsection*{Force Estimation and Localization Metrics}
For force estimation, the maximum value of the predicted force distribution is interpreted as the estimated scalar contact force magnitude, i.e., $\hat{y} = \max_j \hat{\tau}_j$, and is compared with the corresponding ground-truth force magnitude $y$. The mean absolute error (MAE) is defined as
\begin{equation}
\label{eq:mae}
\mathrm{MAE} = \frac{1}{n}\sum_{i=1}^{n}\left| y_i - \hat{y}_i \right|,
\end{equation}
where $n$ denotes the number of evaluation samples.

For contact localization, the predicted contact region is obtained from the index corresponding to the maximum value of the predicted force distribution. The predicted arc-length location is then computed from this index, and the localization error is evaluated using the same MAE definition, computed between the predicted and ground-truth contact positions.
We evaluate collision detection performance under a cross-run generalization setting using the synchronized dataset and the leave-one-test-id split described previously. The classifier outputs a collision probability in the range $(0,1)$, and performance is assessed using the Receiver Operating Characteristic--Area Under the Curve (ROC--AUC), precision, and recall.

\section{Experimental Evaluation and Results}
\begin{figure}[t]
    \centering
    \includegraphics[width = \columnwidth]{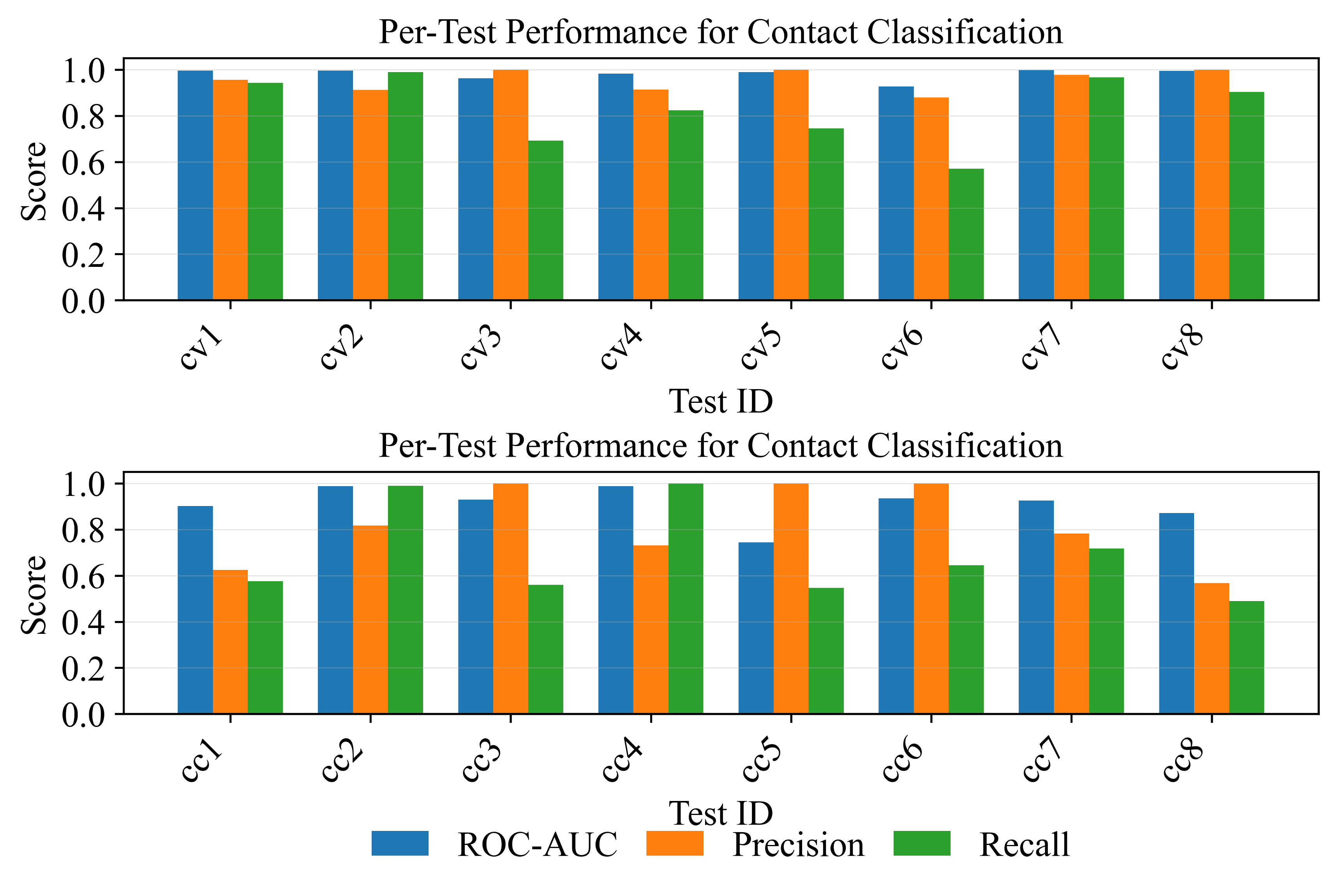}
    \caption{Bar plots showing per-test performance for contact classification, including ROC-AUC, Precision, and Recall scores across different test IDs.}
    \label{fig:contact_classification}
\end{figure}

\begin{figure}[t]
    \centering
    \includegraphics[width = 0.8\columnwidth]{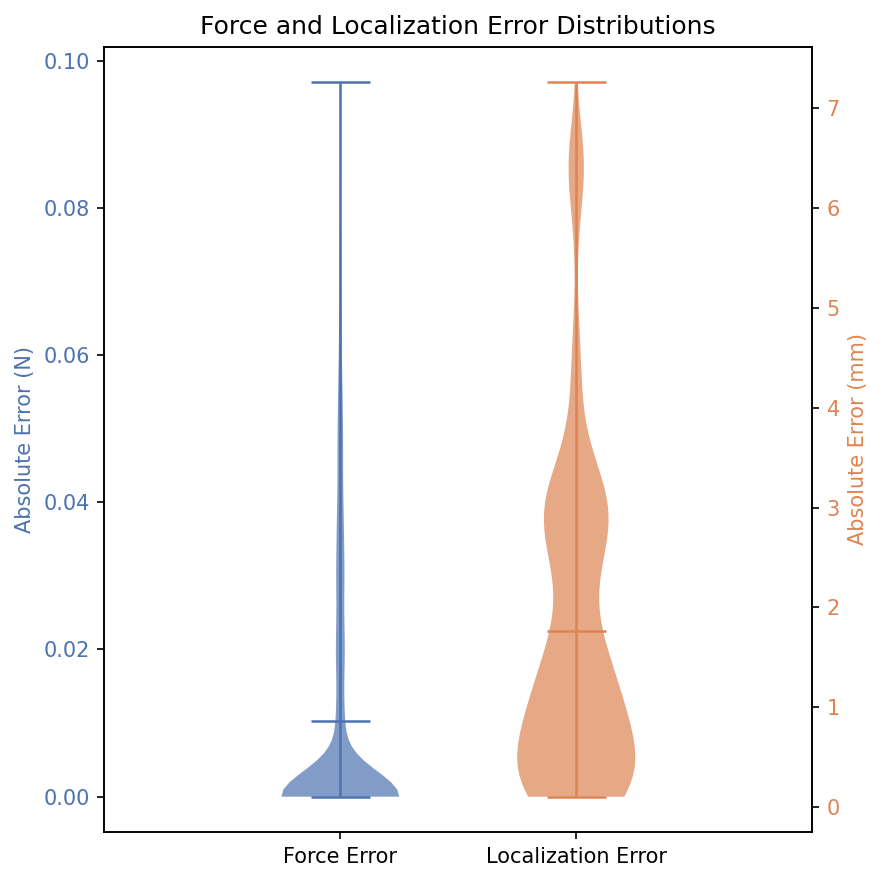}
    \caption{Force and localization absolute error distributions. Mean absolute error (MAE) indicated by horizontal line.}
    \label{fig:error_summary}
\end{figure}

To evaluate the proposed CLF, we used the experimental setup and data collection protocol described previously. The collected dataset was partitioned using a leave-one-test-id cross-validation strategy, where each experimental run was held out entirely for testing while training was performed on the remaining runs, ensuring evaluation on unseen data without temporal leakage. To comprehensively assess the
performance of the CLF model, we then used the aforementioned evaluation metrics.
Figure~\ref{fig:contact_classification} shows the contact classification performance across experimental trials. A complete summary of ROC--AUC, precision, and recall values for all trials is reported in Table~\ref{tab:combined}.  We also evaluated the force estimation and contact localization performance of the proposed CNN--FiLM model using the mean absolute error (MAE) metric defined previously in (\ref{eq:mae}).  A complete summary of classification, force, and localization metrics is provided in Table~\ref{tab:combined}. Figure \ref{fig:error_summary} presents violin plots illustrating the distribution of force and localization of mean absolute errors across the cross-run evaluation. 

\begin{table}[h]
\centering
\caption{Contact Classification Performance Metrics, Force and Localization Errors}
\label{tab:combined}

\begin{tabular}{l c c c c c}
\toprule
Label & \makecell{ROC\\-AUC} & Recall & Precision & \makecell{Force\\MAE (N)} & \makecell{Localization\\MAE (mm)} \\
\midrule
$\mathrm{cv}_{1}$ & 0.995 & 0.943 & 0.957 & 0.013 & 2.936 \\
$\mathrm{cv}_{2}$ & 0.996 & 0.989 & 0.913 & 0.008 & 0.494 \\
$\mathrm{cv}_{3}$ & 0.963 & 0.692 & 1.000 & 0.017 & 0.646 \\
$\mathrm{cv}_{4}$ & 0.983 & 0.824 & 0.913 & 0.008 & 6.477 \\
$\mathrm{cv}_{5}$ & 0.990 & 0.745 & 1.000 & 0.009 & 0.497 \\
$\mathrm{cv}_{6}$ & 0.927 & 0.570 & 0.880 & 0.014 & 0.728 \\
$\mathrm{cv}_{7}$ & 0.999 & 0.966 & 0.977 & 0.006 & 1.298 \\
$\mathrm{cv}_{8}$ & 0.994 & 0.903 & 1.000 & 0.011 & 0.767 \\
\midrule
$\mathrm{cc}_{1}$ & 0.901 & 0.576 & 0.625 & 0.012 & 2.557 \\
$\mathrm{cc}_{2}$ & 0.988 & 0.989 & 0.817 & 0.012 & 2.767 \\
$\mathrm{cc}_{3}$ & 0.929 & 0.560 & 1.000 & 0.019 & 0.594 \\
$\mathrm{cc}_{4}$ & 0.988 & 1.000 & 0.730 & 0.006 & 3.969 \\
$\mathrm{cc}_{5}$ & 0.745 & 0.546 & 1.000 & 0.011 & 2.827 \\
$\mathrm{cc}_{6}$ & 0.935 & 0.645 & 1.000 & 0.014 & 2.416 \\
$\mathrm{cc}_{7}$ & 0.925 & 0.718 & 0.782 & 0.006 & 0.988 \\
$\mathrm{cc}_{8}$ & 0.871 & 0.489 & 0.568 & 0.009 & 2.832 \\
\bottomrule
\end{tabular}
\end{table}


\section{Discussion}

We first evaluate the effectiveness of the CLF framework for \textit{contact classification} under the leave-one-test-id cross-validation setting. As shown in Fig. \ref{fig:contact_classification}, The results demonstrate consistently high ROC--AUC values across most convex surface experiments, with several trials exceeding 0.99. Precision and recall values indicate strong discrimination capability, with limited false positives and missed detections.  Further, as reported in Table~\ref{tab:combined}, convex trials ($\mathrm{cv}_{i}$) achieve consistently high ROC--AUC values, often exceeding 0.99, together with strong precision and recall. Across all convex (cv) and concave (cc) trials, the model achieves low force MAE values, typically below 0.02~N. Several experiments, such as $\mathrm{cv}_{7}$ and $\mathrm{cc}_{7}$, exhibit force MAE values as low as 0.006~N, indicating accurate magnitude reconstruction from distributed strain measurements. These results indicate that distributed OFDR strain profiles provide sufficient discriminative information to reliably separate contact and non-contact states across experimental runs.
Performance in concave trials ($\mathrm{cc}_{i}$) exhibits greater variability, which is consistent with the sensing configuration. Since the optical fiber is embedded on a single side of the NiTi backbone, bending toward the fiber (cv) produces tensile strain with larger amplitude and sharper spatial signatures, whereas bending away from the fiber (cc) produces compressive strain with reduced contrast. This directional sensitivity is mechanically expected and explains the lower recall observed in several concave experiments (e.g., $\mathrm{cc}_{5}$ and $\mathrm{cc}_{8}$). Importantly, even under this asymmetric sensing condition, classification performance remains strong overall, demonstrating that the model learns deformation features that generalize beyond individual runs.

Having established reliable contact detection, as as reported in Table~\ref{tab:combined}, we next considered \textit{force magnitude estimation}. Across all convex and concave trials, the mean absolute error remains below 0.02~N, with several experiments (e.g., $\mathrm{cv}_{7}$ and $\mathrm{cc}_{7}$) achieving MAE values as low as 0.006~N. The consistency of force accuracy across both bending directions suggests that force magnitude is primarily encoded in the overall deformation intensity rather than the exact spatial sign of strain. The distributed nature of OFDR sensing provides dense spatial coverage along the backbone, enabling the CNN--FiLM architecture to extract global load-dependent features that remain stable across experimental conditions. Also, Investigation of Fig. \ref{fig:error_summary} shows that force estimation errors are tightly concentrated near zero, with a median absolute error below 0.01 N and most samples under 0.02 N, while only a few outliers extend toward 0.1 N. The low dispersion and right-skewed distribution indicate stable cross-run generalization. This performance reflects the effectiveness of the Gaussian-encoded supervision and the CNN–FiLM architecture, where predicting a spatial force distribution and extracting its maximum yields a robust scalar force estimate.

Finally, we examined \textit{contact localization}. As summarized  in Table~\ref{tab:combined}, in convex trials, localization errors are typically within sub-millimeter to low-millimeter range (e.g., $\mathrm{cv}_{2}$ and $\mathrm{cv}_{5}$ below 0.5~mm), indicating that the distributed strain field provides strong spatial resolution for identifying the contact region. 
Concave trials generally exhibit slightly higher localization errors compared to convex trials, with several cases exceeding 2~mm. Concave trials generally exhibit slightly higher localization MAE, which again reflects the reduced amplitude and contrast of compressive strain relative to tensile strain in the single-sided fiber configuration. Because localization depends on identifying the peak of the strain perturbation, it is inherently more sensitive to signal contrast than force magnitude estimation.  Also, investigation of Fig. \ref{fig:error_summary} shows that the median localization error is approximately 1–2 mm, with most predictions within 3 mm and a sparse tail extending to ~6–7 mm. Since the manipulator arc length is discretized into 2.65 mm segments (see Fig. \ref{concept}), this value defines the intrinsic spatial resolution of the system; errors below 2.65 mm indicate correct localization within a single segment.  Of note, this 2.65 mm resolution is determined by the physical footprint of the force-gauge contact probe and therefore represents the intrinsic spatial limit of the experimental setup. Larger errors of 6–7 mm correspond to localization within two to three neighboring segments, which remains practically acceptable for interaction-aware tasks such as collision avoidance and contact region identification. Overall, the results demonstrate that the CLF achieves localization accuracy near the physical resolution limit while maintaining robust cross-run generalization.

Overall, the results demonstrate that a single distributed OFDR fiber enables reliable \textit{contact classification}, accurate \textit{force magnitude estimation}, and competitive \textit{contact localization} within the CLF. The directional differences observed between convex and concave conditions are consistent with strain mechanics and reflect the inherent asymmetry of single-sided sensing rather than limitations of the proposed learning architecture. These findings validate that dense distributed strain sensing provides sufficient observability to simultaneously infer interaction occurrence, magnitude, and location without explicit inverse mechanical modeling.
\section{Conclusion and Future Work}

In this paper, we presented a learning-based framework for simultaneous contact detection, localization, and force magnitude estimation in CM using a single distributed OFDR optical fiber. By leveraging continuous strain measurements along the manipulator backbone, the proposed CLF decomposes the interaction inference problem into sequential classification and structured regression stages. Contact detection is formulated as a binary classification task using a Gradient Boosting model trained on distributed strain profiles, while a CNN--FiLM architecture predicts a Gaussian-encoded spatial force distribution to enable joint estimation of contact location and magnitude without explicit inverse mechanical modeling. 

The framework was experimentally validated on a tendon-driven CM instrumented with a single OFDR fiber, with motion-capture-based localization and force guage measurements providing ground truth across multiple contact locations and bending directions. Under leave-one-test-id cross-validation, the proposed approach achieved a mean ROC-AUC value of 0.946, a mean force prediction MAE of 0.011 N, and a mean localization MAE of 2.112 mm, with several trials achieving sub-millimeter localization errors. These results demonstrate that dense distributed strain sensing provides sufficient information to jointly infer interaction occurrence, magnitude, and location within a unified learning formulation. 

Future work will investigate integration of the proposed OFDR-based sensing and CLF framework into clinically relevant continuum platforms such as robotic catheters and flexible endoscopes. In such systems, distributed strain-based interaction awareness can enhance safe navigation in anatomically constrained environments without reliance on external imaging.

\bibliographystyle{IEEEtran}
\bibliography{root}
\end{document}